\documentclass[11pt]{article}

\usepackage[final]{acl}

\usepackage{times}
\usepackage{latexsym}

\usepackage{xspace}
\usepackage{amssymb}
\usepackage{multirow}
\usepackage{amsmath}
\usepackage{booktabs}
\usepackage{xspace}
\usepackage{amssymb}
\usepackage{multirow}
\usepackage{mdframed}
\usepackage{listings}

\newmdenv[
  linewidth=0.5pt,
  innertopmargin=5pt, innerbottommargin=5pt,
  innerleftmargin=6pt, innerrightmargin=6pt,
  skipabove=0pt, skipbelow=0pt,
]{promptframe}

\lstdefinestyle{prompt}{
  basicstyle=\ttfamily\footnotesize,
  breaklines=true, breakatwhitespace=true,
  breakindent=0pt, breakautoindent=true,
  columns=fullflexible, keepspaces=true,
  showstringspaces=false, upquote=true,
  frame=none, numbers=none,
  aboveskip=0pt, belowskip=0pt,
  moredelim=**[is][\bfseries]{@}{@},
}

\usepackage{xcolor}
\usepackage{amssymb}
\usepackage{pifont}
\usepackage{listings}
\usepackage{booktabs}
\usepackage{array}
\usepackage{colortbl}
\usepackage[most]{tcolorbox}

\definecolor{caseframe}{gray}{0.72}
\definecolor{caseback}{gray}{0.975}
\definecolor{casehead}{gray}{0.925}
\definecolor{casemuted}{gray}{0.45}
\definecolor{caseerror}{RGB}{178,34,34}
\definecolor{caseok}{RGB}{34,128,74}

\definecolor{casedsl}{RGB}{0,121,107}
\definecolor{casemodel}{RGB}{204,85,0}

\lstdefinelanguage{LeanCase}{
  morekeywords={def,lemma,theorem,by,have,show,match,with,fun,let,if,then,else,
                calc,namespace,end,open,induction,cases,using},
  sensitive=true,
  morecomment=[l]{--},
  morestring=[b]",
  moredelim=**[is][\color{casedsl}]{@<}{>@},
  moredelim=**[is][\color{casemodel}]{!<}{>!},
}

\lstdefinestyle{leanbase}{
  basicstyle=\ttfamily\scriptsize,
  columns=fullflexible,
  keepspaces=true,
  breaklines=true,
  breakatwhitespace=false,
  postbreak=\mbox{\textcolor{casemuted}{\tiny$\hookrightarrow$}\space},
  showstringspaces=false,
  upquote=true,
  extendedchars=true,
  escapeinside={(*@}{@*)},
  moredelim=**[is][\color{casedsl}]{@<}{>@},
  moredelim=**[is][\color{casemodel}]{!<}{>!},
  aboveskip=0pt,
  belowskip=0pt,
  xleftmargin=0pt,
  numbers=none,
  literate=
    {ℕ}{{$\mathbb{N}$}}1 {ℤ}{{$\mathbb{Z}$}}1
    {≥}{{$\geq$}}1 {≤}{{$\leq$}}1 {¬}{{$\neg$}}1
    {∀}{{$\forall$}}1 {∃}{{$\exists$}}1 {∨}{{$\vee$}}1
    {⊢}{{$\vdash$}}1 {✝}{{$^{\dagger}$}}1 {·}{{$\cdot$}}1
    {←}{{$\leftarrow$}}2 {→}{{$\rightarrow$}}2 {↑}{{$\uparrow$}}1
    {⟨}{{$\langle$}}1 {⟩}{{$\rangle$}}1
    {₀}{{$_0$}}1 {₁}{{$_1$}}1 {₂}{{$_2$}}1 {₃}{{$_3$}}1 {₄}{{$_4$}}1
    {₅}{{$_5$}}1 {₆}{{$_6$}}1 {₇}{{$_7$}}1 {₈}{{$_8$}}1 {₉}{{$_9$}}1 {¹}{{$^1$}}1
    {λ}{{$\lambda$}}1 {↦}{{$\mapsto$}}2 {α}{{$\alpha$}}1 {×}{{$\times$}}1
    {«}{{$\langle\!\langle$}}1 {»}{{$\rangle\!\rangle$}}1,
}

\lstdefinestyle{leancase}{
  style=leanbase,
  language=LeanCase,
  keywordstyle=\bfseries,
  commentstyle=\color{casemuted},
}

\lstdefinestyle{leanplain}{
  style=leanbase,
}

\lstdefinestyle{leanerr}{
  style=leanplain,
  frame=l, framerule=0.9pt, framesep=3pt,
  rulecolor=\color{caseerror!85},
  backgroundcolor=\color{caseerror!5},
  xleftmargin=3.9pt,
}

\lstdefinestyle{leanquote}{
  style=leanbase,
  basicstyle=\sffamily\scriptsize,
  breakatwhitespace=true,
  breakindent=0pt,
  postbreak={},
}

\lstdefinestyle{leancaseapp}{
  style=leancase,
  frame=single,
  framerule=0.3pt,
  rulecolor=\color{caseframe},
  backgroundcolor=\color{caseback},
  framexleftmargin=0pt,
  numberstyle=\tiny\color{casemuted},
  numbersep=3pt,
  aboveskip=2pt,
  belowskip=2pt,
}
\lstdefinestyle{leancaseappnum}{
  style=leancaseapp,
  numbers=left,
  xleftmargin=1.8em,
  framexleftmargin=1.6em,
}
\lstdefinestyle{leanquoteapp}{
  style=leanquote,
  frame=single,
  framerule=0.3pt,
  rulecolor=\color{caseframe},
  backgroundcolor=\color{caseback},
}
\lstdefinestyle{leanplainapp}{
  style=leanplain,
  frame=single,
  framerule=0.3pt,
  rulecolor=\color{caseframe},
  backgroundcolor=\color{caseback},
}

\newtcolorbox{casefigure}[1][]{
  enhanced, breakable=false,
  colback=white, colframe=caseframe, boxrule=0.4pt, arc=1.2pt,
  left=3pt, right=3pt, top=0pt, bottom=1.5pt,
  subtitle style={boxrule=0pt, arc=0pt, colback=casehead, coltext=black,
    fontupper=\scriptsize\sffamily, top=0.4pt, bottom=0.4pt, left=3pt, right=3pt},
  #1
}

\newcommand{\casesubsep}{%
  \par\nointerlineskip\vspace{2pt}%
  \hbox to \linewidth{{\color{caseframe!75}\xleaders\hbox to 2.6pt{\hss\vrule height 0.3pt width 1.3pt\hss}\hfill}}%
  \nointerlineskip\vspace{2.2pt}}

\definecolor{casecol}{gray}{0.93}

\usepackage[T1]{fontenc}

\usepackage[utf8]{inputenc}

\usepackage{microtype}

\usepackage{inconsolata}

\usepackage{graphicx}

\newcommand{\ours}{OEIS-Equiv\xspace}
\title{\ours: Evaluating Code Reasoning in Large Language Models via Equivalence Proofs}

\author{
 \textbf{Yehua Lin\textsuperscript{1}},
 \textbf{Yin Chen\textsuperscript{2,\thanks{Corresponding author.}}}
\\
\\
 \textsuperscript{1}School of Computer Science, South China Normal University, China\\
 \textsuperscript{2}School of Artificial Intelligence, South China Normal University, China
\\
 {
   \texttt{linyehua@m.scnu.edu.cn} \quad
   \texttt{ychen@scnu.edu.cn}
 }
}

\begin{document}
\maketitle

\begin{abstract}
Large language models (LLMs) are increasingly used to turn natural language into code. But does this mean LLMs can reason about code? Existing evaluations can hardly tell, because most use familiar code and check only answers. We therefore turn to code that is valid but unfamiliar. We pair a program found by automated search, written in a Lean 4 domain-specific language, with one that an LLM generated for the same integer sequence. We then ask models to prove the two equivalent. Accordingly, we introduce \ours: 200 such tasks, each with a Lean-verified reference proof. DeepSeek-V4-Pro, which wrote all the LLM-generated programs, proves no pair equivalent, even after reasoning for tens of thousands of tokens per task. No other model proves more than a few. Even on tasks that a single line of automated tactics can prove, most are proved by no model. Thus, for LLMs, writing correct code and reasoning about code are not the same thing.
\end{abstract}

\begin{figure*}[!th]
    \centering
    \includegraphics[width=0.99\textwidth]{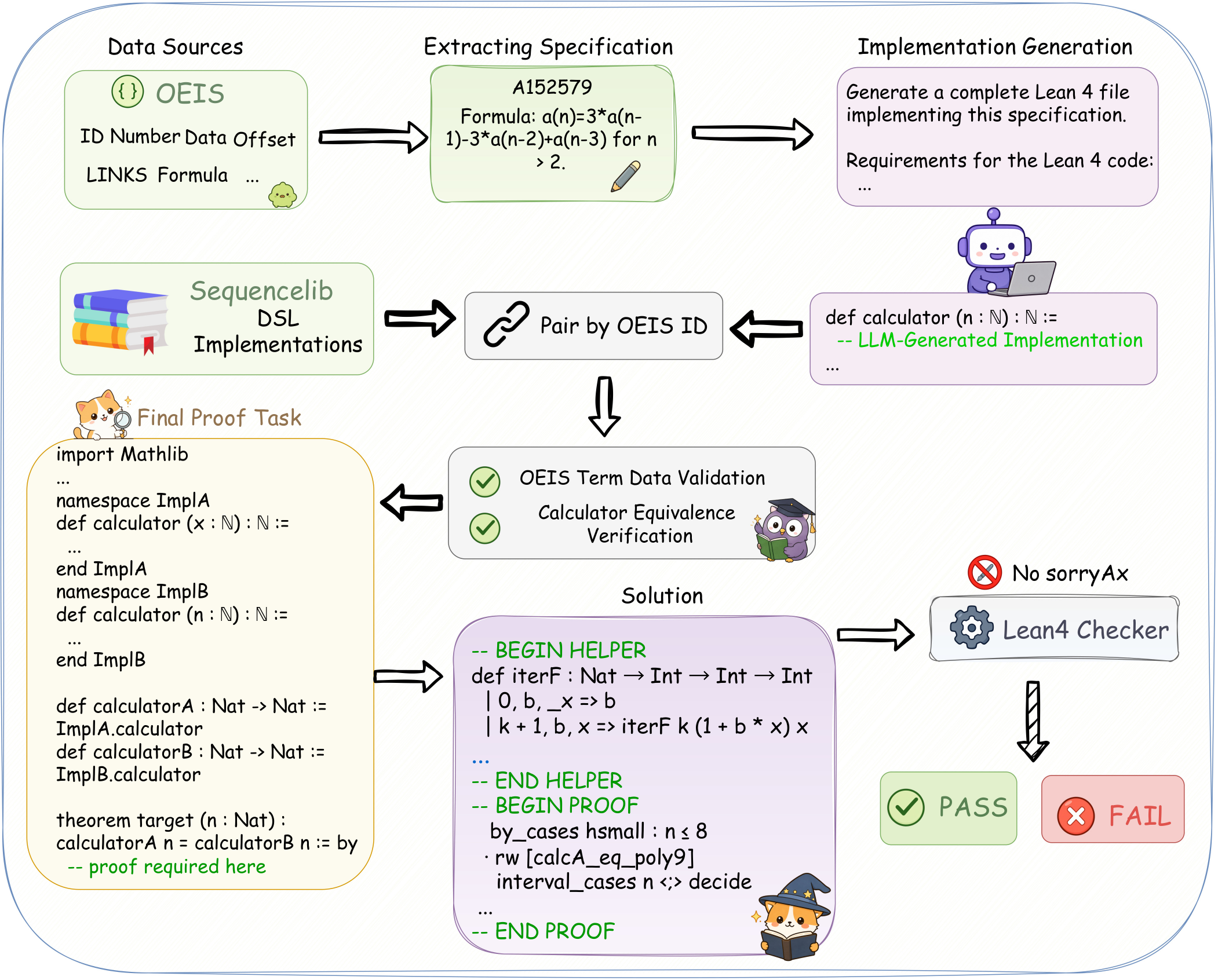}
    \caption{Overview of \ours. }
    \label{fig:overview}
\end{figure*}

\section{Introduction}

Large language models (LLMs) are increasingly used to turn ideas expressed in natural language into code~\cite{CodeGenSurvey,glm5,deepseekai2026deepseekv4highlyefficientmilliontoken}, and current models can usually write correct, executable code in a single attempt, without external tools or compiler feedback. But does this ability mean that LLMs can reason about code~\cite{CodeSense,CodeSemanticsBenchmark}? The same performance may also come from familiar coding patterns seen during training~\cite{FunctionalMemorization,SyntaxWithoutSemantics,SWEBenchIllusion}, much like a student who has memorized the solutions to worked examples: on familiar exercises, such a student may do as well as one who has mastered the methods, and the difference shows only on unfamiliar problems.

Existing evaluations cannot easily tell the two apart. They ask models to predict program outputs~\cite{CRUXEval}, to reason about real-world code~\cite{CodeSense}, or to judge whether two programs are equivalent~\cite{EquiBench,SeqCoBench,CodeSemanticsBenchmark}, and they have two limitations. First, most of them use code in common styles, so a model may succeed by recognizing familiar patterns rather than by following the rules of the language. Second, they check only answers. A yes-or-no verdict may be a lucky guess, and a predicted output may be correct even when the model misreads the program, as long as the misreading does not change the answer. We therefore need two things: code that is valid yet unconventional, conforming to the language's rules while departing from familiar programming idioms, and a task that cannot be completed without correctly interpreting that code.

Programs found by automated search provide such code. \citet{OEISProgramSynthesis} synthesized programs for many sequences in the On-Line Encyclopedia of Integer Sequences (OEIS)~\cite{OEIS}, using a minimal domain-specific language (DSL) built from integer arithmetic, conditionals, and loops, and Sequencelib~\cite{Sequencelib} represents these programs as executable definitions in Lean 4~\cite{Lean}. Found by search and built from a few predefined operators, they look very different from code written by humans or LLMs. As the task, we choose proving program equivalence: a proof must hold for every input and justify every computation step of both programs, so it cannot rest on a wrong reading of either.

Following this idea, we build \ours{} (Figure~\ref{fig:overview}), a benchmark of 200 self-contained Lean 4 tasks. Each task pairs a DSL program with a program written by DeepSeek-V4-Pro~\cite{deepseekai2026deepseekv4highlyefficientmilliontoken} from the OEIS description of the same sequence. The two programs agree on the known terms of the sequence and on an extended range of inputs, and each task comes with a Lean-verified reference proof, so every task is provable. Since the LLM-generated programs were produced in a single attempt without tool feedback, it is reasonable to expect LLMs to prove their equivalence without tool feedback as well. We study the research questions under this setting:
\begin{itemize}
    \item RQ1: Can current LLMs prove that the LLM-generated programs are extensionally equivalent to the DSL programs?
    \item RQ2: Some tasks can be proved directly by Lean's automated proof tactics. Can LLMs prove these tasks?

\end{itemize}

The answers to RQ1 and RQ2 are both largely negative. The most striking contrast comes from DeepSeek-V4-Pro: it wrote all 200 LLM-generated programs in the benchmark but proves none of the equivalences, and no other model proves more than 3 tasks, even though the general-purpose models spend tens of thousands of tokens on a typical task. Even on the 21 tasks that automated proof tactics complete directly, where the proof itself is easy, most are proved by no model. These results suggest that LLMs can write correct code without being able to reason about it.

The main contributions of this paper as follows:
\begin{itemize}
    \item We build \ours{}, a benchmark of 200 self-contained Lean 4 tasks that pair DSL programs from Sequencelib with programs written by DeepSeek-V4-Pro, each with a Lean-verified reference proof.
    \item We evaluate three general-purpose reasoning models and three formal theorem provers, and find that most of the tasks that one-line scripts of standard tactics can prove are proved by no model.
\end{itemize}
\section{Related Work}
\label{sec:related}

\paragraph{OEIS and program synthesis.}
\citet{OEISProgramSynthesis} synthesized programs for many OEIS sequences~\cite{OEIS} in a minimal DSL, and \citet{OEISInductiveBenchmark} turned pairs of these programs into 29{,}687 equivalence conjectures for inductive theorem provers. Both programs in such a pair come from the same DSL, which suits the goal of testing automated provers. \ours{} has a different goal, evaluating code reasoning in LLMs. It pairs each DSL program in Sequencelib~\cite{Sequencelib} with an LLM-written program, checks every pair numerically, and provides a Lean-verified reference proof so that every task is provable.

\paragraph{Program equivalence.}
Recent benchmarks ask LLMs whether two programs are equivalent~\cite{EquiBench,SeqCoBench,CodeSemanticsBenchmark}. With only a yes-or-no answer, a correct verdict may be a lucky guess and reveals little about the model's reasoning. \ours{} instead asks for a proof that the two programs agree on every input, which requires correct reasoning about both programs.

\paragraph{Code reasoning.}
Other work evaluates code reasoning, for example by asking models to predict program inputs and outputs~\cite{CRUXEval} or to reason about real-world code~\cite{CodeSense}. Related studies suggest that strong coding performance may partly come from memorization~\cite{FunctionalMemorization,SWEBenchIllusion} and does not carry over well to an unseen language~\cite{SyntaxWithoutSemantics}. These evaluations mostly check final answers, so reasoning errors that leave the answer unchanged go unnoticed. \ours{} uses valid but unfamiliar code and requires a proof for all inputs, which reveals such errors.
\section{Preliminaries}
\label{sec:prelim}

\subsection{Programs in Lean}

Lean 4 is both a functional programming language and an interactive theorem prover~\cite{Lean}. All sequence implementations in OEIS-Equiv have type $\mathbb{N} \to \mathbb{N}$, but they may compute with integers along the way. In Lean, $\mathbb{N}$ and $\mathbb{Z}$ are \texttt{Nat} and \texttt{Int}. They combine arithmetic, local \texttt{let} bindings, conditionals, pattern matching, and recursion. A proof reasons about such a program by unfolding its definition, for example with \texttt{simp}.

\subsection{Elaboration and Arithmetic}

Lean's elaborator turns surface syntax into fully typed terms. It resolves overloaded operators such as \texttt{-} and \texttt{/} by their expected type and inserts type conversions where needed, so the surface form of an expression does not fully fix its meaning. In an arithmetic expression, Lean chooses one type for the whole expression, taking the expected type into account, and converts each innermost operand to that type. For example, if \texttt{x : Nat} and \texttt{x / 3} is passed to a function that expects an \texttt{Int}, Lean elaborates it as \texttt{(x : Int) / 3}, a division over $\mathbb{Z}$, not as \texttt{((x / 3 : Nat) : Int)}, a division over $\mathbb{N}$ followed by a conversion.

This difference matters because arithmetic behaves differently on the two types. On $\mathbb{N}$, subtraction truncates at zero, while on $\mathbb{Z}$ it can be negative: for $n = 0$, \texttt{((n - 1 : Nat) : Int)} is $0$, whereas \texttt{(n : Int) - 1} is $-1$. On \texttt{Int}, \texttt{/} and \texttt{\%} give the quotient and remainder of Euclidean division, on which the DSL operators \texttt{div} and \texttt{mod} are built; for a divisor $d > 0$, the quotient rounds down and the remainder lies in $[0, d)$. \texttt{Int.toNat} keeps non-negative integers unchanged and maps negative integers to $0$. Finally, \texttt{f <| u + v} applies \texttt{f} to the whole expression on its right, so \texttt{Int.toNat <| u + v} means \texttt{Int.toNat (u + v)}, not \texttt{Int.toNat u + v}. Lean displays such an inserted conversion with the arrow $\uparrow$: \texttt{$\uparrow$x / 3} stands for \texttt{(x : Int) / 3}, whereas \texttt{$\uparrow$(x / 3)} stands for \texttt{((x / 3 : Nat) : Int)}.

\subsection{The DSL}

The language of \citet{OEISProgramSynthesis} consists of arithmetic operators ($+$, $-$, $\times$, \texttt{div}, \texttt{mod}), a conditional $\mathrm{cond}(a, b, c)$ that returns $b$ if $a \le 0$ and $c$ otherwise, two loop operators, and a minimization operator \texttt{compr}. The value $\mathrm{compr}(f, a)$ is the $a$-th (counting from 0) non-negative integer $m$ with $f(m) \le 0$; this operator makes the language Turing-complete~\cite{OEISProgramSynthesis}. Sequencelib~\cite{Sequencelib} defines these operators as Lean functions over $\mathbb{Z}$. Here, \texttt{loop f a b} starts from \texttt{b} and applies \texttt{f} with the counter $1, \ldots, a$ in turn, returning \texttt{b} when $a \le 0$; \texttt{loop2 f g a b c} updates the pair $(b, c)$ to $(f(b, c), g(b, c))$ $a$ times and returns the first component; \texttt{loop'} is the original recursive form of \texttt{loop}; and \texttt{loop2'} is a variant of \texttt{loop2} that keeps the state as a pair and returns the whole pair.

Because \texttt{compr} searches without a bound and may never stop, Sequencelib implements it as \texttt{comprN}, which searches only $0, \ldots, N - 1$ with $N = 1000$ and returns $N$ if the search fails. A program that uses \texttt{comprN} may therefore not be equivalent to a correct implementation for all $n$, so we exclude \texttt{comprN} and all sequences that use it, which leaves ten combinators: \texttt{add}, \texttt{sub}, \texttt{mul}, \texttt{div}, \texttt{mod}, \texttt{cond}, \texttt{loop}, \texttt{loop'}, \texttt{loop2}, and \texttt{loop2'}.

A DSL program takes an input $n : \mathbb{N}$ and returns its result through \texttt{Int.toNat}, while its combinators compute over $\mathbb{Z}$. Hence, depending on the types fixed by elaboration, some parts of a program are evaluated over $\mathbb{N}$ and others over $\mathbb{Z}$.

\subsection{Extensional Equivalence}

For two functions $f, g : \alpha \to \beta$, extensional equivalence is defined as
\begin{equation}
f =_{\mathrm{ext}} g \iff \forall x : \alpha,\; f(x) = g(x).
\label{eq:ext}
\end{equation}
It requires the same output on every input but allows different program structures and intermediate states. In Lean, function extensionality (\texttt{funext}) guarantees that two functions that agree on every input are equal, so \ours states the proof goal directly as pointwise equality.
\section{\ours}

\subsection{Curation}
\paragraph{Extracting Specifications.}
We first extract textual fields containing computational content from each sequence entry in the OEIS data repository. Since computational descriptions in OEIS may span multiple lines, we preserve complete multi-line specification blocks according to their structural markers. For each sequence, this step produces a set of candidate computational descriptions, each of which may correspond to a distinct implementation strategy.

\paragraph{Implementation Generation.}
For each candidate, we use DeepSeek-V4-Pro~\cite{deepseekai2026deepseekv4highlyefficientmilliontoken} to generate the corresponding Lean~4 implementation. The prompt instructs the model to follow the given OEIS description closely and to produce an independent implementation in a standardized format.

\paragraph{Instance Validation.}
We discard implementations with compilation or execution errors and proceed with a two-phase numerical validation. The in-range phase checks the generated code against the given prefix terms from OEIS. The extended-domain phase compares it with the existing Sequencelib implementation on inputs beyond that prefix. Specifically, we require both functions to agree on $\{0,1,\ldots,200\} \cup \{999,1000,1001,4999,5000,5001\}$. Pairs failing any check are removed.

\paragraph{Task Finalization.}
After validation, we strip provenance metadata, including OEIS sequence identifiers and source names, from the implementations presented to models. Placed in separate namespaces to prevent collisions, the functions are exposed purely as generic \(\mathbb{N}\to\mathbb{N}\) types. We then randomly sample 200 retained instances as the final task set. For every final task instance, we construct a corresponding Lean~4 reference proof and verify it under the same criterion applied to LLM outputs.

\begin{table*}[!htb]
\centering
\footnotesize
\setlength{\tabcolsep}{3pt}
\begin{tabular}{@{}lcc@{}}
\toprule
One-line script & Tasks & Year\\
\midrule
\texttt{induction n <;> simp\_all! [\$DEFS] <;> (try zify at *) <;> (grind)} & 17 & 2025\\
\texttt{induction n using Nat.twoStepInduction <;> simp\_all! +arith [\$DEFS] <;> try? +harder} & 12 & 2025\\
\texttt{induction n using Nat.twoStepInduction <;> simp\_all! [\$DEFS] <;> (ring\_nf at * <;> omega)} & 9 & 2024\\
\midrule
\texttt{induction n using Nat.twoStepInduction <;> simp\_all! [\$DEFS] <;> (ring\_nf at * <;> omega)} & 9 & 2024\\
\texttt{simp\_all! [\$DEFS] <;> (try norm\_cast at *; nlinarith)} & 7 & 2022\\
\texttt{induction n using Nat.twoStepInduction <;> simp\_all [\$DEFS] <;> (aesop)} & 2 & 2023\\
\bottomrule
\end{tabular}
\caption{Smallest sets of one-line scripts. Top: one of the smallest sets, whose three scripts together prove all 21 tasks. Bottom: the only smallest set when only tactics available by 2024 are used, whose three scripts together prove the 15 tasks that these tactics can prove. \texttt{\$DEFS}: all definitions of the task. Tasks: number of tasks that a script proves on its own. Year: year in which the newest tactic of the script became available.}
\label{tab:auto}
\end{table*}

\subsection{Statistics}
\label{sec:statistics}

\begin{figure}[!htb]
  \centering
  \includegraphics[width=0.9\columnwidth]{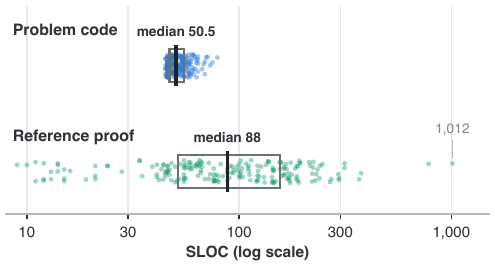}
  \caption{SLOC distributions for task code and reference proofs. Each dot is one task, the box spans the interquartile range, and the heavier vertical line marks the median. Problems are compact and narrowly distributed, whereas reference proofs range from 9 to 1,012 lines.}
  \label{fig:dataset_sloc}
\end{figure}

We measure the size of each task and its reference proof in non-comment, non-blank lines of Lean~4 source code~(SLOC). Task counts include the fixed prelude, both implementations and their supporting definitions, and the target theorem header. Reference-proof counts include all auxiliary definitions and the final proof body. Figure~\ref{fig:dataset_sloc} reports the distributions.

\section{Experiments}
\label{sec:experiments}

\subsection{Evaluation Setup}
\label{sec:setup}

\paragraph{Models.}
We evaluate six models from two groups: the general-purpose reasoning models DeepSeek-V4-Pro~\cite{deepseekai2026deepseekv4highlyefficientmilliontoken}, GLM-5.2~\cite{glm5}, and MiniMax-M3~\cite{lai2026minimaxsparseattention}, and the formal theorem provers DeepSeek-Prover-V2-7B~\cite{dpskProverV2}, Goedel-Prover-V2-8B~\cite{GoedelProverV2}, and Goedel-Prover-V2-32B~\cite{GoedelProverV2}. Each model is required to complete the proof directly, without tool feedback. The general-purpose models run at their highest reasoning effort. The output token limit follows each model’s constraint: 128k tokens for the general-purpose models and 32k tokens for the provers.

\paragraph{Judge.} A proof is accepted only if it passes three checks. First, after comments are removed, it must not contain forbidden commands such as \texttt{sorry}, \texttt{admit}, \texttt{axiom}, \texttt{opaque}, and \texttt{partial}, or metaprogramming keywords such as \texttt{elab}, \texttt{macro}, and \texttt{syntax}. Second, combined with the task into a complete file, it must compile under Lean 4.24.0~\cite{Lean} and Mathlib v4.24.0~\cite{mathlib}. Third, \texttt{\#print axioms} on the target theorem must not report \texttt{sorryAx}.

\paragraph{Automated proof tactics.}
To answer RQ2, we run a series of proof scripts built from automated proof tactics on each task in \ours{} to check whether the task can be proved directly. These scripts combine definition unfolding, optional induction, and closing tactics such as \texttt{omega}, \texttt{aesop}, and \texttt{grind}; the full list is given in Appendix~\ref{app:automation}.

\subsection{Results}
\label{sec:results}

\begin{table}[!htb]
\centering
\setlength{\tabcolsep}{1mm}
\begin{tabular}{lc}
\toprule
Method  & Tasks solved \\
\midrule
DeepSeek-V4-Pro      & 0/200 (0.0\%) \\
GLM-5.2               & 2/200 (1.0\%) \\
MiniMax-M3             & 3/200 (1.5\%) \\
DeepSeek-Prover-V2-7B   & 2/200 (1.0\%) \\
Goedel-Prover-V2-8B    & 0/200 (0.0\%) \\
Goedel-Prover-V2-32B    & 1/200 (0.5\%) \\
\midrule
Automated proof tactics    & 21/200 (10.5\%) \\
\bottomrule
\end{tabular}
\caption{Number of tasks solved by each method on \ours{}. No model proves more than 3 tasks, while 21 tasks can be completed directly by automated proof tactics.}
\label{tab:main_results}
\end{table}

\begin{table*}[!htb]
\centering
\small
\setlength{\tabcolsep}{5pt}
\begin{tabular}{lcccccc}
\toprule
& \multicolumn{2}{c}{DeepSeek-V4-Pro} & \multicolumn{2}{c}{GLM-5.2} & \multicolumn{2}{c}{MiniMax-M3} \\
\cmidrule(lr){2-3}\cmidrule(lr){4-5}\cmidrule(lr){6-7}
& All tasks & 21 tasks & All tasks & 21 tasks & All tasks & 21 tasks \\
\midrule
\multicolumn{7}{l}{Completion tokens, in thousands} \\
\quad Median                & 78.9        & 36.7       & 114.0       & 71.0        & 60.8       & 38.0       \\
\quad 25th--75th percentiles & 57.5--106.0 & 30.7--51.3 & 80.9--131.1 & 61.3--86.9  & 43.9--87.2 & 18.4--52.9 \\
\quad Minimum--maximum      & 20.3--131.1 & 20.3--89.6 & 41.0--131.1 & 41.4--110.7 & 6.5--131.1 & 6.5--64.4  \\
\quad Mean                  & 81.4        & 41.8       & 104.8       & 73.7        & 68.3       & 37.3       \\
\quad Median (non-truncated) & 74.5       & 36.7       & 87.8        & 71.0        & 57.2       & 38.0       \\
\midrule
\multicolumn{7}{l}{Counts} \\
\quad Truncated generations & 19 & 0 & 71 & 0 & 20 & 0 \\
\quad Tasks proved          & 0  & 0 & 2  & 2 & 3  & 3 \\
\bottomrule
\end{tabular}
\caption{Completion tokens and outcomes for the three general-purpose models. All tasks: the 200 tasks of \ours{}. 21 tasks: the tasks that one-line scripts can prove. 25th--75th percentiles: the interval between the 25th and 75th percentiles of completion length. Median (non-truncated): the median over generations that end before the output limit of 131,072 tokens.}
\label{tab:token_usage}
\end{table*}

Table~\ref{tab:main_results} reports the results. All models struggle with the equivalence proofs (RQ1). The most striking contrast comes from DeepSeek-V4-Pro: it wrote all 200 LLM-generated programs in \ours{}, yet it fails to prove any of them equivalent to the corresponding DSL program. No other model proves more than 3 tasks. This holds even for the general-purpose models, which run at their highest reasoning effort with an output limit of 128k tokens, four times that of the provers. Most of their generations stop on their own before this limit, and the rest reach it without finding a proof (Section~\ref{sec:tokens}). For RQ2, 21 tasks have a one-line proof built from automated proof tactics, yet 13 of them are proved by no model. These results suggest that, compared with their ease in writing code, LLMs fall short in reasoning about it.

\section{Further Analysis}
\label{sec:analysis}

\subsection{Automated Proof Tactics}
\label{sec:automation}

We first analyze the automated proof tactics. Three one-line scripts are enough to prove all 21 tasks (Table~\ref{tab:auto}, top). The three scripts follow the same pattern: they first set up an induction on $n$ if needed, then unfold all definitions of the task with \texttt{simp\_all!}, and finally pass the remaining goal to a closing tactic. The first script uses ordinary induction, moves the goal from $\mathbb{N}$ to $\mathbb{Z}$ with \texttt{zify}, and completes the proof with \texttt{grind}, which follows the approach of SMT solvers and combines equality reasoning, case splits, and arithmetic solving. This script alone proves 17 tasks. The other two scripts use two-step induction, which assumes the claim for $n$ and $n+1$ when proving it for $n+2$. One of them ends with \texttt{try?}, which searches among many tactics, including \texttt{grind}. This script proves the three tasks that the other scripts miss. The other first normalizes polynomials with \texttt{ring\_nf} and then ends with \texttt{omega}, a decision procedure for linear arithmetic over $\mathbb{N}$ and $\mathbb{Z}$ that also handles truncated subtraction and division by constants.

However, some of these tactics are recent: \texttt{try?} appeared only in March 2025, and \texttt{grind} became an official Lean tactic only in August 2025. DeepSeek-Prover-V2 and Goedel-Prover-V2 were both released in 2025 and are unlikely to have encountered these tactics during training. We therefore repeat the search using only tactics that were available by the end of 2024. These older tactics still prove 15 of the 21 tasks, and only one set of three scripts can prove all 15 tasks together (Table~\ref{tab:auto}, bottom). Among them, the two-step script with \texttt{omega} appears again and proves 9 tasks. The second script needs no induction: after unfolding the definitions, it handles the conversions between $\mathbb{N}$ and $\mathbb{Z}$ with \texttt{norm\_cast} and completes the proof with \texttt{nlinarith}, which adds products of hypotheses to linear arithmetic. The third script uses two-step induction and ends with \texttt{aesop}, a general proof search tactic.

The proof itself is easy on these 21 tasks. Each of them has a one-line proof, and 15 of them can be proved with tactics that predate all evaluated models. Yet every task proved by a model is among these 15, and 7 of the 15 are proved by no model. On these tasks, the obstacle is therefore unlikely to be the proof itself and more likely the reasoning about the two programs. Choosing such a line still requires reasoning about the programs, namely whether an induction is needed, whether it needs one step or two, and which closing tactic suits the arithmetic left after unfolding. The closing tactic then handles truncated subtraction, integer division, and conversions between $\mathbb{N}$ and $\mathbb{Z}$ strictly as Lean defines them.

\begin{figure}[!htb]
\centering
\includegraphics[width=0.8\columnwidth]{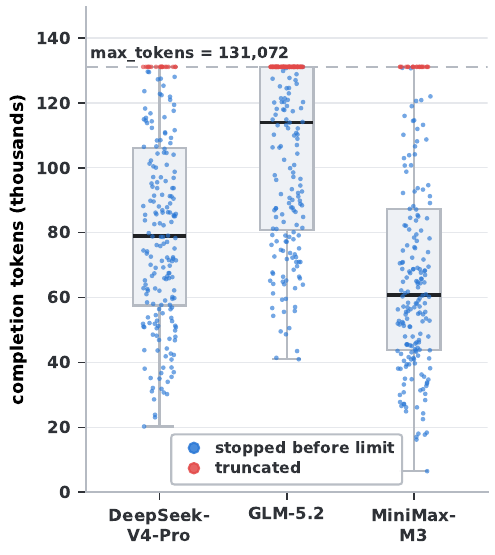}
\caption{Completion tokens per generation for the three general-purpose models. Each point is one generation. The dashed line marks the output limit of 131,072 tokens, and red points mark generations that reach it.}
\label{fig:token_usage}
\end{figure}

\subsection{Token Usage}
\label{sec:tokens}

Section~\ref{sec:automation} shows that 21 tasks can be proved by one-line scripts. We now look at how long the general-purpose models reason before they give a proof. As described in Section~\ref{sec:setup}, these models run at their highest reasoning effort with an output limit of 128k tokens. Figure~\ref{fig:token_usage} and Table~\ref{tab:token_usage} report the number of completion tokens in each generation.

The models reason at length. The median generation length ranges from 60.8k to 114.0k tokens across models. Yet most generations end on their own before the limit (Table~\ref{tab:token_usage}), so a higher limit would not have changed them, and nearly all of them still fail. Truncation also does not explain the differences between models. GLM-5.2 reaches the limit most often but proves 2 tasks, whereas DeepSeek-V4-Pro reaches it least often and proves none.

On the 21 tasks that one-line scripts can prove, the limit plays no role at all. No generation reaches it, and the median generation length ranges from 36.7k to 71.0k tokens. The models reason for tens of thousands of tokens on tasks that one line of tactics can close, but most of these tasks are proved by no model.

The models thus have ample room to reason, yet their reasoning rarely leads to a correct proof. 

\section{Conclusion}
\label{sec:conclusion}

Writing correct code is now easy for LLMs. Reasoning about it is a different matter. \ours{} separates the two by asking models to prove that a familiar program and an unfamiliar one compute the same thing. Current models largely fail, even when a short script of standard tactics would do. We hope this work encourages evaluating code LLMs beyond their outputs, and building models that read a program by what it means rather than by how it looks.

\bibliography{custom}

\appendix

\section{Automation Setting}
\label{app:automation}

We test each task against a catalog of 481 one-line script templates in 21 families (Table~\ref{tab:families}). In a template, \texttt{\$DEFS} stands for all definitions in the task file and \texttt{\$NAT} for the variable $n$ of the target theorem. Most templates optionally start with an induction on $n$ (ordinary, two-step, strong, or strong with a case split), unfold the definitions, optionally normalize the goal, and end with a closing tactic such as \texttt{omega}, \texttt{aesop}, \texttt{nlinarith}, or \texttt{grind}; the others rely on search tactics such as \texttt{exact?}, \texttt{try?}, and \texttt{hint}. Each script serves as the whole proof body and is checked under the same criterion as the model proofs. A run that exceeds the time or heartbeat limit counts as a failure.

\begin{table*}[!htp]
\centering
\footnotesize
\frenchspacing
\begin{tabular}{@{}>{\raggedright\arraybackslash}p{4.1cm}r>{\raggedright\arraybackslash}p{9.4cm}@{}}
\toprule
Family & Templates & Example\\
\midrule
Single tactic & 22 & \texttt{omega}\\
Unfold, then close & 27 & \texttt{simp\_all [\$DEFS] <;> (omega)}\\
Unfold listed definitions only, then close & 12 & \texttt{simp\_all only [\$DEFS] <;> (omega)}\\
Unfold with a discharger, then close & 36 & \texttt{simp\_all (disch := omega) [\$DEFS] <;> (aesop)}\\
Unfold and normalize, then close & 29 & \texttt{simp\_all! [\$DEFS] <;> (try norm\_cast at *; nlinarith)}\\
Move to $\mathbb{Q}$, then close & 16 & \texttt{qify at * <;> (nlinarith)}\\
Clear denominators, then close & 4 & \texttt{field\_simp (discharger := \ldots) at * <;> (ring)}\\
Induction, unfold, then close & 64 & \texttt{induction \$NAT <;> simp\_all [\$DEFS] <;> (omega)}\\
Induction, unfold and normalize, then close & 68 & \texttt{induction \$NAT <;> simp\_all! [\$DEFS] <;> (ring\_nf at * <;> omega)}\\
Linear and nonlinear arithmetic & 24 & \texttt{simp\_all! +arith [\$DEFS] <;> nlinarith!}\\
Congruence, then close & 48 & \texttt{(congr! 1) <;> (omega)}\\
Extensionality, then close & 6 & \texttt{ext1 <;> simp\_all! [\$DEFS] <;> (omega)}\\
\texttt{grind} with definitions & 6 & \texttt{grind [\$DEFS]}\\
\texttt{grind} with other settings & 19 & \texttt{grind +splitImp [\$DEFS]}\\
\texttt{aesop} with other settings & 25 & \texttt{aesop (config := \ldots)}\\
Library search & 12 & \texttt{try? +harder}\\
\texttt{try?} with other options & 23 & \texttt{try? -only}\\
Other search tactics & 6 & \texttt{hint}\\
Rewrite search & 18 & \texttt{rw? <;> (omega)}\\
Backward search & 6 & \texttt{solve\_by\_elim (config := \ldots)}\\
Other algebra & 10 & \texttt{abel\_nf at * <;> grind}\\
\midrule
Total & 481 & \\
\bottomrule
\end{tabular}
\caption{Families of one-line script templates, with one example each. Most families also include variants with other simplification, normalization, or closing steps.}
\label{tab:families}
\end{table*}

\section{Environment}
\label{app:environment}

All proof checking runs on a single machine with an Intel Xeon Gold 6330 @ 2.00\,GHz (100 CPUs) and 176\,GiB of RAM, under Ubuntu 22.04.4 LTS with kernel 6.8.0-111. Proof checking uses Lean 4.24.0 with Mathlib v4.24.0, against a pre-built Mathlib cache so that no task pays elaboration cost for the library itself. Each candidate is compiled in its own temporary directory with a 180-second timeout, and candidates are checked in parallel with a bounded worker pool. The harness that drives generation, extraction, integrity screening and checking is written in Python 3.12.

\begin{table*}[!htb]
\centering
\small
\begin{tabular}{@{}p{0.38\textwidth}p{0.56\textwidth}@{}}
\toprule
Model & Model ID \\
\midrule
DeepSeek-V4-Pro       & \texttt{deepseek-ai/}\allowbreak\texttt{DeepSeek-V4-Pro} \\
GLM-5.2               & \texttt{zai-org/}\allowbreak\texttt{GLM-5.2} \\
MiniMax-M3            & \texttt{MiniMaxAI/}\allowbreak\texttt{MiniMax-M3} \\
DeepSeek-Prover-V2-7B & \texttt{deepseek-ai/}\allowbreak\texttt{DeepSeek-Prover-V2-7B} \\
Goedel-Prover-V2-8B   & \texttt{Goedel-LM/}\allowbreak\texttt{Goedel-Prover-V2-8B} \\
Goedel-Prover-V2-32B  & \texttt{Goedel-LM/}\allowbreak\texttt{Goedel-Prover-V2-32B} \\
\bottomrule
\end{tabular}
\caption{LLMs used in this work.}
\label{tab:model_ids}
\end{table*}

\section{Models}
\label{app:models}

Table~\ref{tab:model_ids} lists the LLMs that appear in this work together with the exact version we used.

\section{Prompt}
\label{app:prompts}

The prompt for evaluation is given in Figure~\ref{fig:standard}, with \{prelude\} replaced by the task file.

\begin{figure*}[!htb]
\begin{promptframe}
\begin{lstlisting}[style=prompt]
You are given a Lean4 file defining two functions `calculatorA` and `calculatorB`. Your task is to prove their universal equality for an arbitrary natural number.

The checker will synthesize the final theorem as:

```lean
theorem target (n : Nat) : _root_.calculatorA n = _root_.calculatorB n := by
  -- your proof body here
```

You may introduce auxiliary helper definitions or lemmas, but only inside the `-- BEGIN HELPER` / `-- END HELPER` block. Put the final proof body, without a leading `by`, inside the `-- BEGIN PROOF` / `-- END PROOF` block, and indent it by two spaces as if it were placed immediately after `:= by`.

Do not return a full Lean file, include any `import` lines, modify existing definitions, or introduce proof placeholders or new assumptions such as `sorry`, `admit`, `axiom`, or `constant`.

Return exactly one Lean code block in this format:

```lean
-- BEGIN HELPER
-- optional helper definitions / lemmas here
-- END HELPER

-- BEGIN PROOF
  -- proof body here, no leading `by`
-- END PROOF
```

Here is the immutable Lean task prelude:

```lean
@{prelude}@
```

You must prove this theorem:

```lean
theorem target (n : Nat) : calculatorA n = calculatorB n := by
  -- proof required here
```

The proof will be checked using Lean 4.24.0 and mathlib v4.24.0.
\end{lstlisting}
\end{promptframe}
\caption{Prompt for evaluation.}
\label{fig:standard}
\end{figure*}

\section{The On-Line Encyclopedia of Integer Sequences~(OEIS)}
\label{app:oeis}

The OEIS is a database of integer sequences maintained by The OEIS Foundation Inc. Neil J. A. Sloane began collecting sequences on file cards in 1964 as a graduate student, published them as \emph{A Handbook of Integer Sequences} (2{,}372 sequences) in 1973 and, with Simon Plouffe, as \emph{The Encyclopedia of Integer Sequences} (5{,}487 sequences) in 1995, and put the collection online in 1996. The intellectual property was transferred to the Foundation in 2009, and submissions have since been reviewed by a panel of volunteer editors. As of July 2026 the database holds 397{,}811 entries.

\begin{figure*}[!htb]
\centering
\includegraphics[width=\textwidth]{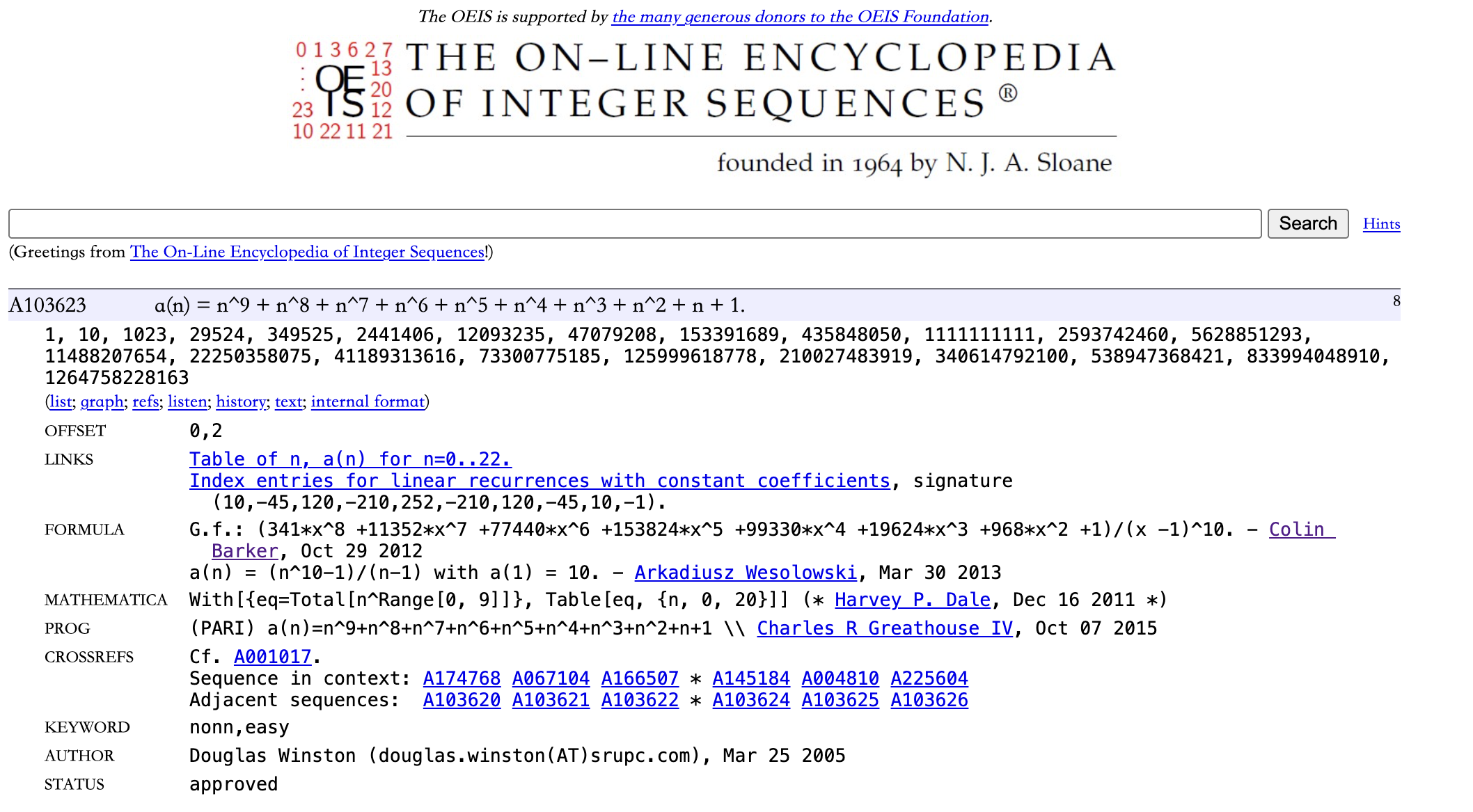}
\caption{An OEIS entry. Below the identifier, the name and the leading terms, the entry collects references, formulas, cross-references and programs. The Formula field often describes the same sequence in different computational terms.}
\label{fig:oeisexample}
\end{figure*}

Each entry carries a permanent identifier of the form A\emph{nnnnnn}, so that a sequence has essentially one place to be looked up. An entry gives some or all of the following: the leading terms, a name, the offset of the first term, comments, references and links, formulas, cross-references to related entries, programs in systems such as Maple, Mathematica and PARI, the contributor, and the full edit history. Figure~\ref{fig:oeisexample} shows one entry as it appears on the website.

The OEIS is made available under the Creative Commons Attribution-ShareAlike 4.0 license (CC BY-SA 4.0), which permits copying and adapting the material, provided that the OEIS is credited and adapted material is shared under the same license. Sequencelib~\cite{Sequencelib} is released under the same license.

\section{Example}
\label{app:case}

This section gives the complete tasks of two examples, verbatim~(Figures~\ref{fig:app-task1} and~\ref{fig:app-task2}).

\clearpage
\begin{figure}[tbp]
\begin{lstlisting}[style=leancaseapp, basicstyle=\ttfamily\fontsize{7}{7.5}\selectfont]
import Mathlib

namespace Synth

open Function

def add (a b : Int) : Int := a + b
def sub (a b : Int) : Int := a - b
def mul (a b : Int) : Int := a * b
def div (a b : Int) : Int := a / b
def mod (a : Int) (b : Int) : Int := a % b
def cond (a b c : Int) : Int := if a ≤ 0 then b else c

def loop' (f : Int → Int → Int) (a : Int) (b : Int) : Int :=
  match a with
  | .ofNat 0 => b
  | .ofNat (n + 1) => f (loop' f n b) (n + 1)
  | .negSucc _ => b
termination_by Int.natAbs a

def loop (f : Int → Int → Int) (n : Int) (b : Int) : Int :=
  match n with
  | .ofNat m => List.range' 1 m |>.map (↑·) |>.foldl f b
  | .negSucc _ => b

def loop2 (f g : Int → Int → Int) (a : Int) (b c : Int) : Int :=
  match a with
  | .ofNat 0 => b
  | .ofNat (n + 1) => loop2 f g n (f b c) (g b c)
  | .negSucc _ => b
termination_by Int.natAbs a

def loop2' (f g : Int × Int → Int) (a : Int) (b : Int × Int) : Int × Int :=
  match a with
  | .ofNat 0 => b
  | .ofNat (n + 1) => loop2' f g n (f b, g b)
  | .negSucc _ => b
termination_by Int.natAbs a

end Synth

namespace ImplA

open Synth

open Synth

def calculator (x : ℕ) : ℕ :=
  Int.toNat <| loop (λ (x _y) ↦ (x * x) - x) 1 (x / 3) + x

end ImplA

namespace ImplB

open Synth

def calculator (n : Nat) : Nat :=
  let a := n / 3
  let b := n % 3
  a * (a + 2) + b

end ImplB

def calculatorA : Nat -> Nat := ImplA.calculator
def calculatorB : Nat -> Nat := ImplB.calculator

theorem target (n : Nat) : calculatorA n = calculatorB n := by
  -- proof required here
\end{lstlisting}
\caption{Case~1, the complete task given to the models.}
\label{fig:app-task1}
\end{figure}

\begin{figure}[tbp]
\begin{lstlisting}[style=leancaseapp, basicstyle=\ttfamily\fontsize{7}{7.5}\selectfont]
import Mathlib

namespace Synth

open Function

def add (a b : Int) : Int := a + b
def sub (a b : Int) : Int := a - b
def mul (a b : Int) : Int := a * b
def div (a b : Int) : Int := a / b
def mod (a : Int) (b : Int) : Int := a % b
def cond (a b c : Int) : Int := if a ≤ 0 then b else c

def loop' (f : Int → Int → Int) (a : Int) (b : Int) : Int :=
  match a with
  | .ofNat 0 => b
  | .ofNat (n + 1) => f (loop' f n b) (n + 1)
  | .negSucc _ => b
termination_by Int.natAbs a

def loop (f : Int → Int → Int) (n : Int) (b : Int) : Int :=
  match n with
  | .ofNat m => List.range' 1 m |>.map (↑·) |>.foldl f b
  | .negSucc _ => b

def loop2 (f g : Int → Int → Int) (a : Int) (b c : Int) : Int :=
  match a with
  | .ofNat 0 => b
  | .ofNat (n + 1) => loop2 f g n (f b c) (g b c)
  | .negSucc _ => b
termination_by Int.natAbs a

def loop2' (f g : Int × Int → Int) (a : Int) (b : Int × Int) : Int × Int :=
  match a with
  | .ofNat 0 => b
  | .ofNat (n + 1) => loop2' f g n (f b, g b)
  | .negSucc _ => b
termination_by Int.natAbs a

end Synth

namespace ImplA

open Synth

open Synth

def calculator (n : ℕ) : ℕ :=
  let x := n - 1
  Int.toNat <| 1 + if x ≤ 0 then 1 else 2

end ImplA

namespace ImplB

open Synth

def hyper_op (op : Nat) (a : Nat) : Nat :=
  match op with
  | 0 => 2
  | 1 => a + 1
  | _ => a

def chain (k : Nat) : Nat :=
  match k with
  | 0 => hyper_op 0 1
  | k+1 => hyper_op (k+1) (chain k)

def calculator (n : Nat) : Nat :=
  match n with
  | 0 => 2
  | k+1 => chain k

end ImplB

def calculatorA : Nat -> Nat := ImplA.calculator
def calculatorB : Nat -> Nat := ImplB.calculator

theorem target (n : Nat) : calculatorA n = calculatorB n := by
  -- proof required here
\end{lstlisting}
\caption{Case~2, the complete task given to the models.}
\label{fig:app-task2}
\end{figure}

\end{document}